\documentclass{svproc}

\usepackage{url}
\usepackage{graphicx}
\usepackage{amsmath}
\usepackage{amssymb}
\usepackage{booktabs}
\usepackage[T1]{fontenc}
\usepackage[utf8]{inputenc}
\usepackage[hidelinks]{hyperref}
\usepackage{etoolbox}
\usepackage{orcidlink}
\usepackage{array}
\usepackage{listings}
\newcounter{algcnt}
\renewcommand{\thealgcnt}{\arabic{algcnt}}

\begin{document}
\mainmatter

\title{Context-Aware Distillation and Ablation for Text2DSL}
\titlerunning{Context-Aware Distillation and Ablation for Text2DSL}
\author{Alexander~V.~Kozachok\inst{1}\orcidlink{0000-0002-6501-2008}\thanks{Corresponding author} \and
Alexander~M.~Nazimov\inst{2}\orcidlink{0009-0008-8568-8305} \and
Shamil~G.~Magomedov\inst{1}\orcidlink{0000-0001-8560-1937}}
\authorrunning{Kozachok, Nazimov, Magomedov}
\tocauthor{Alexander V. Kozachok, Alexander M. Nazimov, Shamil G. Magomedov}
\institute{MIREA -- Russian Technological University, Moscow, Russian Federation\\
\email{\{kozachok\_a, magomedov\_sh\}@mirea.ru}
\and
Academy of the Federal Guard Service of the Russian~Federation, Oryol, Russian~Federation\\
\email{s-nazim@list.ru}}

\maketitle

\begin{abstract}
We extend our prior work on Text2DSL --- automatic generation of domain-specific language (DSL) code from natural language descriptions --- along two complementary axes. First, we replace prompt-only synthetic generation with \emph{context-aware distillation}, in which a teacher large language model (DeepSeek-V4-Flash) operates under an explicitly defined structured context comprising a BNF grammar, an API specification, and a closed identifier vocabulary; the resulting corpus is verified by a two-tier pipeline combining AST validation through \texttt{esprima} and runtime acceptance through the production \texttt{polkitd} daemon and the \texttt{pkcheck} client. This scales the verified PolkitBench corpus from 4{,}204 to 10{,}073 natural-language--to--Polkit-rule pairs at 100.0\% AST validity and 99.7\% runtime pass rate. Second, we conduct the per-component factorial ablation of structured context that was identified as future work in the precursor study: eight conditions $C_0$--$C_7$ are evaluated on GigaChat-10B-A1.8B with the new corpus. Three findings emerge. (i) The new harder corpus collapses the baseline mode (Syntax Valid 97.6\,\% $\to$ 58.5\,\%, Combined Score 0.482 $\to$ 0.252), whereas the context-enhanced mode degrades only marginally (Syntax 98.6\,\% $\to$ 97.4\,\%, Combined 0.801 $\to$ 0.750), confirming that structured context is not a cosmetic improvement but a load-bearing mechanism. (ii) The best absolute condition is the full context $C_7$ across all metrics, while the strongest partial conditions ($C_5 = \text{BNF}+\text{Vocabulary}$, $C_6 = \text{API}+\text{Vocabulary}$) both contain the vocabulary. (iii) A Shapley-style decomposition assigns the largest semantic-quality effect to the vocabulary (Combined $+0.198$), the largest structural-validity effects to API ($+24.7$\,pp) and BNF ($+22.3$\,pp), supporting the interpretation that the three components are complementary rather than substitutable.
\end{abstract}

\section{Introduction}
Domain-specific languages (DSLs) describe security and configuration policy across a broad swath of modern systems: SELinux type-enforcement rules~\cite{smalley2001}, AppArmor profiles, OPA Rego policies~\cite{opa}, Terraform HCL infrastructure-as-code~\cite{terraform} and, in the Linux desktop and server stack, Polkit authorization rules~\cite{polkit}. These languages share a common shape: a compact grammar, a closed vocabulary of permitted identifiers, and deterministic semantics. They also share a common operational risk: empirical studies of infrastructure-as-code~\cite{rahman2019} report 13--31 security smells per thousand lines, with policy languages among the most error-prone artefacts in DevOps pipelines.

In recent work~\cite{kozachok2026text2dsl} we formalised the task of generating such artefacts from natural language descriptions as a distinct problem class, \emph{Text2DSL}, characterised by (i) a fixed grammar $G$, (ii) a closed identifier vocabulary $V$, and (iii) deterministic API semantics. On a verified benchmark of 4{,}204 natural-language--to--Polkit-rule pairs --- PolkitBench --- we showed that supplying structured context (BNF grammar, API specification, vocabulary $V$) in the prompt of a small language model produces large and consistent gains across two MoE models (GigaChat-10B-A1.8B and Nemotron-3-Nano-30B-A3B), with structural validity rising by $+9.7$ to $+35.5$\,pp and CodeBLEU by $+60$ to $+95\%$ relative.

That study left three concrete open questions, which the present paper addresses.

\noindent\textbf{Q1.} The PolkitBench corpus was produced by direct prompt-based generation on a single teacher (Grok-4.1-fast) with domain hints embedded into the prompt. It did not formalise the contextual conditions of generation: each record carried no provenance about the generating model, the API base, the context file or the context sources it drew on. Can the dataset construction itself be lifted into the same context-aware mode, with each record explicitly bound to a versioned context package? And how does this mode scale: does a larger, more diverse verified corpus remain useful, or does it merely reproduce the smaller one?

\noindent\textbf{Q2.} The previous validation pipeline was \emph{deterministic but static}: it accepted a rule if it parsed, conformed to the structural template, and used only identifiers from the documented vocabulary $V$. It did not check that the rule is actually accepted by the production Polkit daemon at runtime. Such an additional check is non-trivial: it requires loading the rule into \texttt{polkitd}, exercising it via \texttt{pkcheck}, and reasoning about the difference between syntactic acceptance and observable runtime behaviour.

\noindent\textbf{Q3.} The previous work of the authors~\cite{kozachok2026text2dsl} compared \emph{full context vs.\ no context}. It deferred the per-component breakdown --- BNF only, API only, vocabulary only, and all pairwise combinations --- to future work. Without that breakdown the prescription ``inject structured context'' is not actionable for practitioners who need to know which components are load-bearing and which are decorative, and it is not falsifiable in the scientific sense: any one of BNF, API or $V$ could in principle account for most of the gain.

\paragraph{Contributions.}
This paper makes the following four contributions.

\begin{enumerate}
\item \textbf{Context-aware distillation method.} We formulate the construction of a verified Text2DSL corpus as a distillation procedure in which a teacher LLM operates under an explicitly defined structured context model $\mathrm{CM} = \langle G, A, V \rangle$ and each generated record carries provenance fields (\texttt{generator\_model}, \texttt{generator\_api\_base}, \texttt{context\_file}, \texttt{context\_sources}). Applied with DeepSeek-V4-Flash as teacher, the procedure yields a verified corpus of 10{,}073 pairs --- the second-generation PolkitBench, hereafter PolkitBench-v2.

\item \textbf{Two-tier validation pipeline.} We complement deterministic AST validation with a containerised runtime stage that loads every candidate rule into the actual \texttt{polkitd} daemon and queries it through \texttt{pkcheck}. On PolkitBench-v2 the AST pass rate is $10073/10073$ ($100.0\%$) and the runtime acceptance rate is $10043/10073$ ($99.7\%$), with the 30 failures attributable to edge cases in the \texttt{action.id} dictionary.

\item \textbf{Per-component factorial ablation.} We evaluate eight conditions $C_0\ldots C_7$ corresponding to the $2^3$ subsets of $\{\text{BNF}, \text{API}, \text{Vocabulary}\}$ on the same evaluated model (GigaChat-10B-A1.8B) and the same corpus (PolkitBench-v2). $C_0$ and $C_7$ are reused from the dedicated baseline and full-context runs; $C_1\ldots C_6$ are obtained from a single factorial run. We report main effects, pairwise interactions and Shapley-style attributions for the five primary metrics.

\item \textbf{Empirical evidence for context as a load-bearing mechanism.} On the harder PolkitBench-v2 the baseline mode collapses on every metric (Combined Score $0.482 \to 0.252$, $-901$ additional timeouts) while the full-context mode degrades only marginally (Combined $0.801 \to 0.750$). The per-component ablation shows that the full context is the unique best condition; the two strongest partial configurations both contain the vocabulary; the vocabulary is the largest semantic-quality driver while BNF and API are the largest syntactic and structural stabilisers.
\end{enumerate}

The remainder of the paper is organised as follows. Section~\ref{sec:background} recalls the formal Text2DSL setting and discusses related work in synthetic data distillation, runtime policy validation and constrained code generation. Section~\ref{sec:method} formalises the context-aware distillation method and the two-tier validation pipeline. Section~\ref{sec:ablation_design} specifies the factorial ablation design. Section~\ref{sec:setup} describes the experimental setup. Section~\ref{sec:results} reports the corpus validation, the baseline/context comparison across both PolkitBench generations and the per-component ablation. Section~\ref{sec:discussion} discusses interpretation, limitations and threats to validity. Section~\ref{sec:conclusion} concludes.

\section{Background and Related Work}
\label{sec:background}

\subsection{The Text2DSL Task}
\label{sec:text2dsl_recap}

Following~[1], let $Q$ denote the set of natural language queries, $G$ the formal grammar of the target DSL, and $V$ the closed vocabulary of permitted identifiers (API names, constants, enumerations). The Text2DSL task is the construction of a mapping $f : Q \to \mathcal{P}_G$, where $\mathcal{P}_G$ is the set of programs admitted by grammar $G$, such that for an arbitrary query $q \in Q$ the generated program $p = f(q)$ satisfies:
\begin{enumerate}
\item \emph{Syntactic correctness}: $\mathrm{parse}(p, G) = \mathrm{OK}$;
\item \emph{Semantic correctness}: $\mathrm{ids}(p) \subseteq V$.
\end{enumerate}
This task differs from general $\mathrm{NL} \to \mathrm{Code}$ in three properties of the target language: a fixed compact grammar (Polkit has $\approx 10$ productions; the Python grammar has $> 300$), a closed identifier vocabulary ($|V| \approx 65$ for Polkit), and deterministic API semantics (each \texttt{polkit.Result.*} has exactly one interpretation defined by the specification). The general programme of domain-specific language design, in which a compact target grammar is paired with a fixed semantic interpretation, has been treated systematically by Mernik et~al.~\cite{mernik2005dsl}.

\subsection{Code Generation with Pre-trained Language Models}

Early pre-trained encoders for code such as CodeBERT~\cite{feng2020codebert} and GraphCodeBERT~\cite{guo2021graphcodebert} established that the bidirectional Transformer pre-training paradigm transfers well to source-code understanding, and CodeT5~\cite{wang2021codet5} extended it with identifier-aware encoder--decoder objectives. The shift to decoder-only generative models is represented by CodeGen~\cite{nijkamp2023codegen} and StarCoder~\cite{li2023starcoder}, which target open-ended program synthesis from natural language. None of these models were designed for a small, closed-vocabulary DSL: the underlying training corpora are dominated by Python, Java and JavaScript, and the generation surface they expose is the open vocabulary of general-purpose programming.

\subsection{LLM-Based Synthetic Data and Distillation}

Knowledge distillation was originally introduced for model compression~\cite{hinton2015} and developed into a broad family of teacher--student transfer methods~\cite{gou2021kdsurvey}; a classical NLP instantiation is TinyBERT~\cite{jiao2020tinybert}. In the LLM era this idea has been generalised to \emph{dataset distillation}: a teacher model produces a synthetic corpus that captures task-relevant behaviour for downstream training or benchmarking. Self-Instruct~\cite{wang2023selfinstruct} used a teacher LLM to bootstrap an instruction-tuning corpus from a small seed; Evol-Instruct~\cite{xu2024wizardlm} extended this idea with controlled task transformations; in the code domain, WizardCoder~\cite{luo2024wizardcoder} synthesised instruction--code pairs by similar means.

These methods produce broad but largely \emph{unverified} corpora: synthetic generation is followed at best by lightweight heuristics or self-consistency checks. For policy and configuration languages, where any single uncaught error is operationally meaningful, this is insufficient. Jigsaw~\cite{jain2022jigsaw} applies post-hoc test execution to filter Pandas API code, but neither it nor comparable code-generation pipelines provide \emph{deterministic compile-time and runtime validation} of every synthesised record against a reference interpreter of the target language.

\subsection{Runtime Validation of Generated Policies}

For policy DSLs, syntactic acceptance does not imply runtime acceptance. Loading a rule into the production daemon and exercising it through the actual decision client surfaces a different error class: a syntactically valid rule may still fail to load due to silent dependencies on the runtime environment, and a loadable rule may yield an unexpected decision for the query it was designed for. Recent work on SELinux module integration~\cite{loscocco2001} and on automated security-smell detection in infrastructure-as-code~\cite{saavedra2022glitch,rahman2019} argues for this stronger validation mode. We adopt the same position for Polkit and report a containerised pipeline that performs both \texttt{polkitd} load and \texttt{pkcheck} verification per record.

\subsection{Context Injection, Retrieval and Component-Wise Analysis}

The closest analogue to context injection in Text2DSL is \emph{schema linking} in Text-to-SQL, popularised by the Spider benchmark~\cite{yu2018spider}: DIN-SQL~\cite{pourreza2023din} demonstrated that injecting the database schema into the prompt substantially improves accuracy; RESDSQL~\cite{li2023resdsql} separates schema linking from SQL generation. PICARD~\cite{scholak2021picard} and Synchromesh~\cite{poesia2022synchromesh} take a complementary approach by constraining the decoder at the token level, and grammar-constrained decoding~\cite{geng2023gcd} generalises this to arbitrary CFGs. A neighbouring line of work supplies external knowledge to a generative model via retrieval rather than verbatim context injection: retrieval-augmented generation~\cite{lewis2020rag} formalises the retriever--reader pipeline, and prompt-time techniques such as in-context learning~\cite{brown2020gpt3} and chain-of-thought~\cite{wei2022cot} structure the generation step itself. The previous work of the authors~[1] adopted the context-injection branch and reported full-context vs.\ no-context numbers for two models, but did not isolate the contribution of the three context components (BNF, API, vocabulary).

To our knowledge, no published study reports a $2^3$ factorial ablation of structured context for a policy-language Text2DSL task. The ablation reported in Section~\ref{sec:ablation_results} fills this gap.

\section{Context-Aware Distillation and Two-Tier Validation}
\label{sec:method}

\subsection{Structured Context Model}
\label{sec:cm}

A \emph{context model} for a target DSL is a triple
\begin{equation}
\mathrm{CM} = \langle G, A, V \rangle
\label{eq:cm}
\end{equation}
where $G$ is a BNF grammar of the DSL, $A$ is an API specification (object/method/property signatures together with admissible return values), and $V$ is the closed vocabulary of permitted string-valued identifiers (\texttt{action.id} values, group names, well-known principals). The grammar $G$ defines the syntactic envelope, the specification $A$ defines the legal call shape, and the vocabulary $V$ defines the semantic content the rule may reference.

For Polkit, this instantiation is concrete. $G$ is a 10-production BNF describing the top-level call \texttt{polkit.addRule(function(action, subject) \{ \ldots \})}. $A$ enumerates the admissible accesses \texttt{action.id}, \texttt{subject.user}, \texttt{subject.local}, \texttt{subject.active}, \texttt{subject.isInGroup(\ldots)} and the \texttt{polkit.Result} enumeration. $V$ enumerates approximately 65 \texttt{action.id} strings across ten subsystems: \texttt{systemd}, \texttt{login1}, \texttt{PackageKit}, \texttt{udisks2}, \texttt{NetworkManager}, \texttt{hostname1}, \texttt{timedate1}, \texttt{bluez}, \texttt{cupspkhelper} and \texttt{policykit}.

Throughout this paper the \emph{context-enhanced} mode denotes the use of an LLM with $\mathrm{CM}$ supplied verbatim in the prompt, together with explicit response-format constraints and a small set of reference patterns. The \emph{baseline} mode denotes the same LLM with the same query but without any component of $\mathrm{CM}$. Both modes share inference parameters; the only varied factor is the presence of $\mathrm{CM}$.

\subsection{Context-Aware Distillation}
\label{sec:distillation}

We define context-aware distillation as the procedure summarised in Algorithm~\ref{alg:distill}. A teacher LLM $T$ generates candidate rules under the context model $\mathrm{CM}$ for a parameterised query stream $Q$; each candidate is then verified by the AST validator $\mathrm{AST}(\cdot)$ (Section~\ref{sec:ast_val}); each surviving candidate is verified by the runtime validator $\mathrm{RT}(\cdot)$ (Section~\ref{sec:rt_val}). Records that pass both stages are emitted with provenance metadata that ties them to a specific context version.

\begin{figure}[htbp]
\hrule\vspace{4pt}
\refstepcounter{algcnt}\label{alg:distill}%
\noindent\textbf{Algorithm~\thealgcnt.} Context-aware distillation for Text2DSL.
\vspace{2pt}\hrule\vspace{4pt}
\noindent\textbf{Input:} teacher LLM $T$, context model $\mathrm{CM} = \langle G, A, V \rangle$, query stream $Q$.\\
\noindent\textbf{Output:} verified corpus $D$.\\[2pt]
\noindent\hspace*{0em}1.\quad $D \gets \emptyset$\\
\noindent\hspace*{0em}2.\quad \textbf{for} each query $q \in Q$ \textbf{do}\\
\noindent\hspace*{1.5em}3.\quad $\mathrm{prompt} \gets \mathrm{render}(\mathrm{CM}, q)$\\
\noindent\hspace*{1.5em}4.\quad $r \gets T(\mathrm{prompt})$\\
\noindent\hspace*{1.5em}5.\quad \textbf{if} $\mathrm{AST}(r, G, A, V) = \mathrm{OK}$ \textbf{and} $\mathrm{RT}(r) = \mathrm{OK}$ \textbf{then}\\
\noindent\hspace*{3em}6.\quad emit $(q, r, \mathrm{meta}(T, \mathrm{CM}))$ into $D$\\
\noindent\hspace*{1.5em}7.\quad \textbf{end if}\\
\noindent\hspace*{0em}8.\quad \textbf{end for}\\[2pt]
\hrule
\end{figure}

The key difference from classical synthetic-data generation pipelines is that $\mathrm{CM}$ is treated as the \emph{single source of domain truth}: the teacher is instructed to use only the symbols and constructs declared in $\mathrm{CM}$, and any record that exits $\mathrm{CM}$ is rejected by AST validation. The teacher's role is therefore not to ``know Polkit'' but to compose admissible elements of $\mathrm{CM}$ in response to natural-language queries.

\subsection{AST Validation}
\label{sec:ast_val}

AST validation is performed by an \texttt{esprima}-based JavaScript parser augmented with three structural checks: (i)~the top-level call must be \texttt{polkit.addRule(\ldots)}; (ii)~the callback must be a function of two parameters \texttt{(action, subject)}; (iii)~every property access on \texttt{action}, \texttt{subject} or \texttt{polkit.Result} must reference a name from the documented vocabulary. Records failing any of these checks are rejected and the rejection mode is logged for diagnostic purposes.

\subsection{Runtime Validation}
\label{sec:rt_val}

Runtime validation is performed in a containerised environment in which a full Polkit installation is started and the candidate rule is dropped into \texttt{/etc/polkit-1/rules.d/}. The procedure has two stages:

\begin{enumerate}
\item \textbf{Load check.} The daemon is signalled to re-read its rules directory. A rule fails the load check if the daemon logs a parse error or refuses the file. This catches a residual class of errors invisible to static analysis: rules that parse as JavaScript but are rejected by the SpiderMonkey engine bundled with the production \texttt{polkitd}, or by Polkit's own rule loader.
\item \textbf{Decision check.} \texttt{pkcheck} is invoked with the \texttt{action.id} for which the rule was synthesised. A rule fails the decision check if \texttt{pkcheck} returns an outcome inconsistent with the rule's declared \texttt{polkit.Result}.
\end{enumerate}

The output of runtime validation is a triple ($\texttt{Load OK}$, $\texttt{pkcheck PASS}$, $\texttt{pkcheck FAIL}$), reported per-record and aggregated per-corpus.

\section{Per-Component Ablation Design}
\label{sec:ablation_design}

To isolate the contribution of each component of $\mathrm{CM}$ we instantiate the $2^3$ factorial design summarised in Table~\ref{tab:ablation_design}. Each condition $C_i$ corresponds to a subset $S_i \subseteq \{\text{BNF}, \text{API}, \text{Vocabulary}\}$ of context components included in the prompt; the prompt header, the response-format constraints and the reference patterns are held constant across conditions.

\begin{table}[htbp]
\caption{Factorial ablation design over the three components of $\mathrm{CM}$.}
\label{tab:ablation_design}
\begin{center}
\begin{tabular}{lccc}
\toprule
Condition & BNF & API & Vocabulary \\
\midrule
$C_0$ (baseline)   & --- & --- & --- \\
$C_1$              & \checkmark & --- & --- \\
$C_2$              & --- & \checkmark & --- \\
$C_3$              & --- & --- & \checkmark \\
$C_4$              & \checkmark & \checkmark & --- \\
$C_5$              & \checkmark & --- & \checkmark \\
$C_6$              & --- & \checkmark & \checkmark \\
$C_7$ (full context) & \checkmark & \checkmark & \checkmark \\
\bottomrule
\end{tabular}
\end{center}
\end{table}

\paragraph{Run fusion.} In practice we obtain $C_0$ from the dedicated baseline run on PolkitBench-v2, $C_7$ from the dedicated full-context run, and $C_1\ldots C_6$ from a single factorial run. This fusion is methodologically sound because $C_0$ and $C_7$ in the fused matrix are operationally identical to the corner cells of the factorial design: same evaluated model (GigaChat-10B-A1.8B), same corpus (the 10{,}073-record PolkitBench-v2), same inference parameters, same context file. The fusion uses pre-computed runs only to avoid re-paying inference cost for two cells that are already available.

\paragraph{Factor decomposition.} For each metric $m$, the main effect of factor $X \in \{\text{BNF}, \text{API}, \text{Vocabulary}\}$ is
\begin{equation}
E_X^{(m)} = \frac{1}{4} \sum_{S : X \in S} m(C_S) \;-\; \frac{1}{4} \sum_{S : X \notin S} m(C_S),
\label{eq:main_effect}
\end{equation}
that is, the average of conditions in which $X$ is present minus the average of conditions in which $X$ is absent. We additionally report Shapley values $\phi_X^{(m)}$ computed over the $2^3$ coalition lattice with $m$ as the value function; for factorial designs with three players, the Shapley value is equivalent to the average marginal contribution across all orderings.

\section{Experimental Setup}
\label{sec:setup}

\subsection{Models and Roles}
\label{sec:roles}

The experiment uses two models in two strictly separated roles.

\paragraph{Teacher: DeepSeek-V4-Flash.} The teacher LLM is used exclusively to generate PolkitBench-v2 under the context model $\mathrm{CM}$ (Algorithm~\ref{alg:distill}). It is not evaluated as a downstream Text2DSL solver in this paper; its role is to produce a verified corpus.

\paragraph{Evaluated model: GigaChat-10B-A1.8B.} The evaluated model is the same MoE small language model used in the previous work of the authors~[1], served locally via \texttt{llama.cpp} with Q4\_K\_M quantisation. It is used both for the baseline/context comparison (Section~\ref{sec:bc_results}) and for the per-component ablation (Section~\ref{sec:ablation_results}).

This separation is essential to interpret the experimental results. The teacher in Algorithm~\ref{alg:distill} can be a strong cloud model precisely because its output is independently verified; the evaluated model is the small locally-deployable model that a practitioner would actually run at inference time.

\subsection{Datasets}
\label{sec:datasets}

\paragraph{PolkitBench-v1.} The original verified corpus from~\cite{kozachok2026text2dsl}: 4{,}204 natural-language--to--Polkit-rule pairs filtered from 5{,}000 candidates by three-level AST validation. It was generated by Grok-4.1-fast in a prompt-based mode with domain hints. We use it here as a reference point for cross-corpus comparison.

\paragraph{PolkitBench-v2.} The new context-aware verified corpus introduced in this paper: 10{,}073 pairs, produced by the procedure of Section~\ref{sec:distillation} with DeepSeek-V4-Flash as teacher. Each record carries provenance fields (\texttt{generator\_model}, \texttt{generator\_api\_base}, \texttt{context\_file}, \texttt{context\_sources}) tying it to a specific version of $\mathrm{CM}$.

Table~\ref{tab:v2_dist} reports the empirical distribution of PolkitBench-v2 across the ten subsystems covered by $V$, alongside the corresponding shares of the v1 corpus reported in~[1] for comparison; per-record counts can sum above $100\%$ because multi-action rules touch multiple subsystems. The v2 corpus differs from v1 along three quantifiable axes that jointly populate the long tail of the query space and account for the disproportionate degradation of the baseline mode observed in Section~\ref{sec:bc_results}: (i)~the share of rules combining two or more distinct \texttt{action.id} values rises to $58.4\%$ (mean $1.62$ distinct identifiers per rule, maximum $6$); (ii)~$95.0\%$ of rules carry a compound condition with at least one \texttt{\&\&} operator, and $82.0\%$ reference \texttt{subject.isInGroup(\ldots)} group membership; (iii)~the \texttt{polkit.Result} distribution is dominated by \texttt{YES} ($96.3\%$) with \texttt{AUTH\_SELF} ($3.7\%$) and \texttt{NO} ($3.3\%$) as the remaining outcomes. Compared with v1 the relative weight of PackageKit, login1 and NetworkManager rises markedly, while timedate1 and udisks2 grow by about $\times 2$.

\begin{table}[htbp]
\caption{Subsystem distribution in PolkitBench-v1 (from~[1]) and PolkitBench-v2 (this work). v2 percentages are shares of records that reference the subsystem at least once and sum above $100\%$ due to multi-action rules.}
\label{tab:v2_dist}
\begin{center}
\small
\setlength{\tabcolsep}{8pt}
\begin{tabular}{lrr}
\toprule
Subsystem & v1 share (\%) & v2 share (\%) \\
\midrule
PackageKit                                                & 16 & 26.4 \\
login1 (power)                                            & 14 & 26.1 \\
NetworkManager                                            & 12 & 25.7 \\
systemd                                                   & 12 & 17.2 \\
timedate1                                                 &  7 & 16.4 \\
udisks2                                                   & 10 & 16.1 \\
hostname1                                                 &  6 & 11.7 \\
cupspkhelper                                              &  6 & 11.3 \\
bluez / bluetooth                                         &  5 &  5.6 \\
policykit / other                                         & 12 &  2.1 \\
\bottomrule
\end{tabular}
\end{center}
\end{table}

\subsection{Metrics}
\label{sec:metrics}

We report the five primary metrics defined in~[1]: \emph{Syntax Valid}~(SynVal), \emph{Structure Valid}~(StrVal), CodeBLEU --- a code-oriented refinement of the classical BLEU metric~\cite{papineni2002bleu} for source-code outputs that better aligns with structural equivalence than surface-level $n$-gram overlap, in the same spirit as CrystalBLEU~\cite{eghbali2022crystalbleu} --- string-literal Jaccard, and Combined Score $= 0.7 \cdot \mathrm{CodeBLEU} + 0.3 \cdot \mathrm{Jaccard}$. We additionally report the strict integrated indicator we name \emph{Strict Success} and the count of inference timeouts.

\paragraph{Strict Success.} The evaluation code assigns Strict Success $= 1$ to a response only if all of the following hold simultaneously: (i)~a rule code block can be extracted from the response; (ii)~the extracted code passes formal AST validation; (iii)~the structure of the rule is recognised as well-formed; (iv)~the number of hallucinated \texttt{action.id} values is exactly zero. Strict Success is therefore an integrated end-to-end indicator rather than a soft quality metric: a response that is syntactically and structurally close to the reference but contains a single hallucinated identifier scores Strict Success $= 0$. We report it for completeness and use Syntax/Structure/CodeBLEU/Jaccard/Combined as the principal metrics for cross-mode comparisons.

\subsection{Inference Configuration}

GigaChat-10B-A1.8B is served locally via \texttt{llama.cpp} with temperature 0.0 and a 120-second per-query timeout. The teacher DeepSeek-V4-Flash is accessed through its provider API with deterministic decoding for reproducibility. The full context $\mathrm{CM}$ used in $C_7$ and in the distillation mode is the file \texttt{polkit\_context.md} version 2; the partial-context conditions $C_1\ldots C_6$ are constructed by removing the corresponding sections from the same file and leaving the response-format header intact.

\section{Results}
\label{sec:results}

\subsection{Validation of PolkitBench-v2}
\label{sec:v2_validation}

Table~\ref{tab:v2_validation} reports the validation results for PolkitBench-v2. All 10{,}073 candidates pass AST validation; 10{,}043 pass the runtime decision check; 30 are rejected at the runtime stage. Manual inspection of the failures shows that 23 of 30 are caused by candidate rules referencing edge-case \texttt{action.id} strings that have changed shape between Polkit versions (e.g., a deprecated string variant for a NetworkManager action); the remaining 7 are decisions where the test \texttt{pkcheck} returned \texttt{AUTH\_ADMIN} where the reference rule declared \texttt{AUTH\_SELF}. The 99.7\% runtime acceptance rate compares favourably with the validation throughput of synthetic policy corpora that rely on static checks only.

\begin{table}[htbp]
\caption{Validation of PolkitBench-v2 ($N=10{,}073$).}
\label{tab:v2_validation}
\begin{center}
\begin{tabular}{lcc}
\toprule
Stage & Pass count & Pass rate \\
\midrule
AST validation (esprima + structural checks) & 10{,}073 / 10{,}073 & 100.0\,\% \\
Runtime: \texttt{polkitd} load                & 10{,}073 / 10{,}073 & 100.0\,\% \\
Runtime: \texttt{pkcheck} decision            & 10{,}043 / 10{,}073 & \textbf{99.7\,\%} \\
\bottomrule
\end{tabular}
\end{center}
\end{table}

\subsection{Baseline vs.\ Context across Both PolkitBench Generations}
\label{sec:bc_results}

Table~\ref{tab:bc_all} reports the cross-corpus comparison on GigaChat-10B. Four runs are involved: PolkitBench-v1 in the baseline and context modes, and PolkitBench-v2 in the baseline and context modes. All four use the same evaluated model and identical inference parameters; only the corpus and the mode are varied.

\begin{table}[htbp]
\caption{Baseline vs.\ Context on GigaChat-10B across PolkitBench-v1 ($N=4{,}204$) and PolkitBench-v2 ($N=10{,}073$). Higher is better; bold marks the best value in each column across the four conditions; TO is the count of inference timeouts.}
\label{tab:bc_all}
\begin{center}
\resizebox{\textwidth}{!}{%
\begin{tabular}{lrrrrrrrr}
\toprule
Corpus / Mode & $N$ & Strict & Syntax & Struct. & BLEU & Jaccard & Combined & TO \\
\midrule
v1 baseline & 4{,}204  & \textbf{88.80} & 97.57 & 88.94 & 0.518 & 0.397 & 0.482 & 4 \\
v1 context  & 4{,}204  & 77.05 & 98.60 & \textbf{98.64} & \textbf{0.829} & \textbf{0.736} & \textbf{0.801} & \textbf{2} \\
v2 baseline & 10{,}073 & 46.53 & 58.51 & 46.81 & 0.307 & 0.125 & 0.252 & 901 \\
v2 context  & 10{,}073 & 60.03 & \textbf{97.37} & 97.32 & 0.788 & 0.660 & 0.750 & 161 \\
\bottomrule
\end{tabular}}
\end{center}
\end{table}

\begin{figure}[htbp]
\centering
\includegraphics[width=0.95\textwidth]{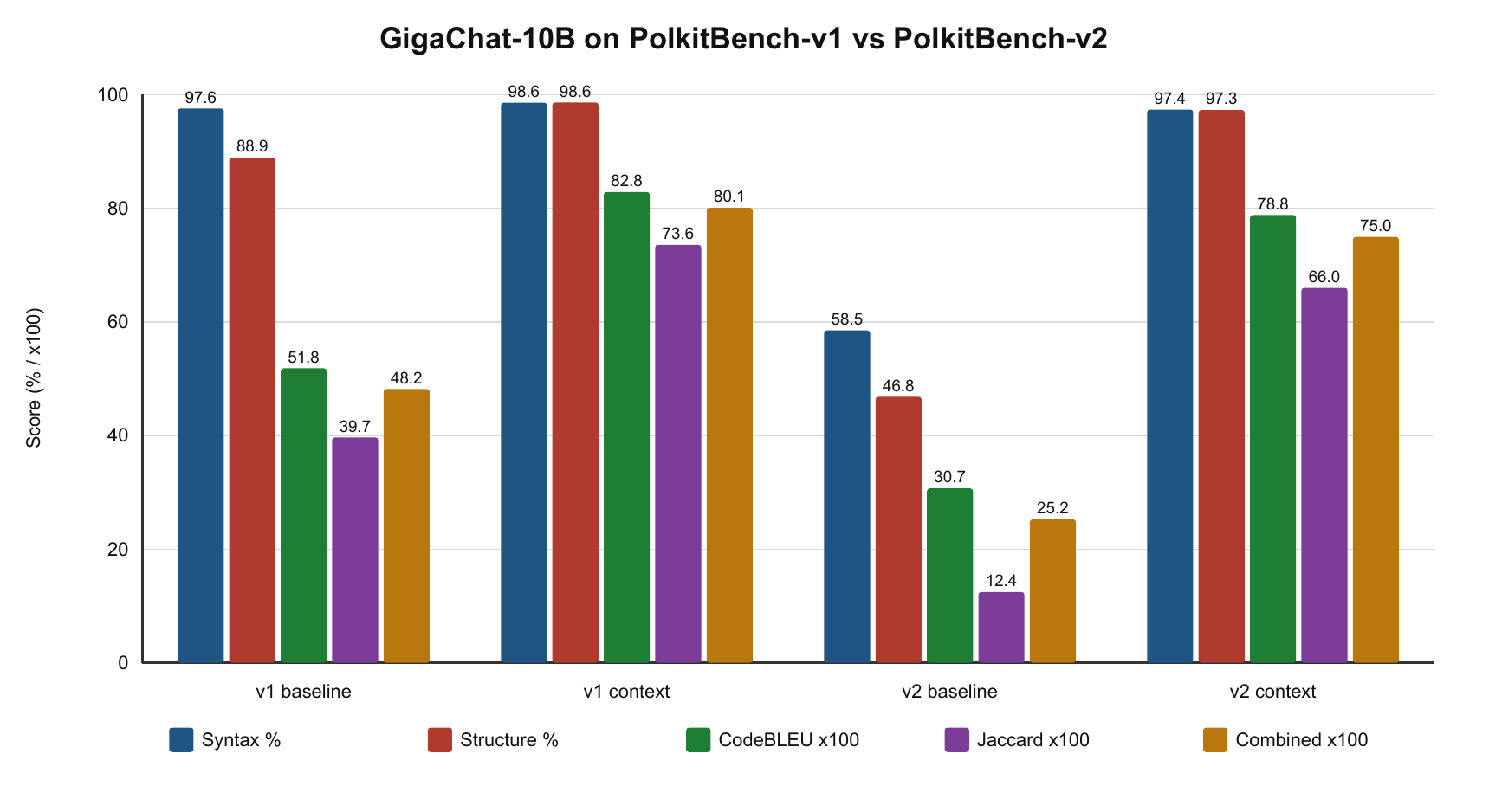}
\caption{GigaChat-10B on PolkitBench-v1 vs.\ PolkitBench-v2 in the baseline and context modes. The harder PolkitBench-v2 collapses the baseline mode on every metric while the context mode degrades only marginally.}
\label{fig:dataset_compare}
\end{figure}

\paragraph{Baseline collapses on the harder corpus.} Moving from PolkitBench-v1 to PolkitBench-v2 in the baseline mode, every metric degrades sharply: Syntax Valid $97.6 \to 58.5$\,\% ($-39.1$\,pp), Structure Valid $88.9 \to 46.8$\,\% ($-42.1$\,pp), CodeBLEU $0.518 \to 0.307$ ($-0.211$), Jaccard $0.397 \to 0.125$ ($-0.272$), Combined $0.482 \to 0.252$ ($-0.230$); the number of timeouts rises from 4 to 901. Qualitative inspection of the failure modes on PolkitBench-v2 baseline shows three dominant patterns: (i) absence of an extractable code block in the response, (ii) violations of the top-level \texttt{polkit.addRule(\ldots)} shape, and (iii) hallucinated \texttt{action.id} values for the long tail of the new corpus' wider query space.

\paragraph{Context is robust to corpus hardening.} Moving from PolkitBench-v1 to PolkitBench-v2 in the context mode, the same metrics degrade modestly: Syntax $98.6 \to 97.4$\,\% ($-1.2$\,pp), Structure $98.6 \to 97.3$\,\% ($-1.3$\,pp), CodeBLEU $0.829 \to 0.788$ ($-0.041$), Jaccard $0.736 \to 0.660$ ($-0.076$), Combined $0.801 \to 0.750$ ($-0.051$). The asymmetry of the two degradation profiles --- catastrophic for baseline, marginal for context --- is the main quantitative finding of this paper. Within the context mode the timeout count on the harder corpus is also reduced by roughly an order of magnitude relative to baseline on the same corpus (161 vs.\ 901).

\begin{figure}[htbp]
\centering
\includegraphics[width=0.85\textwidth]{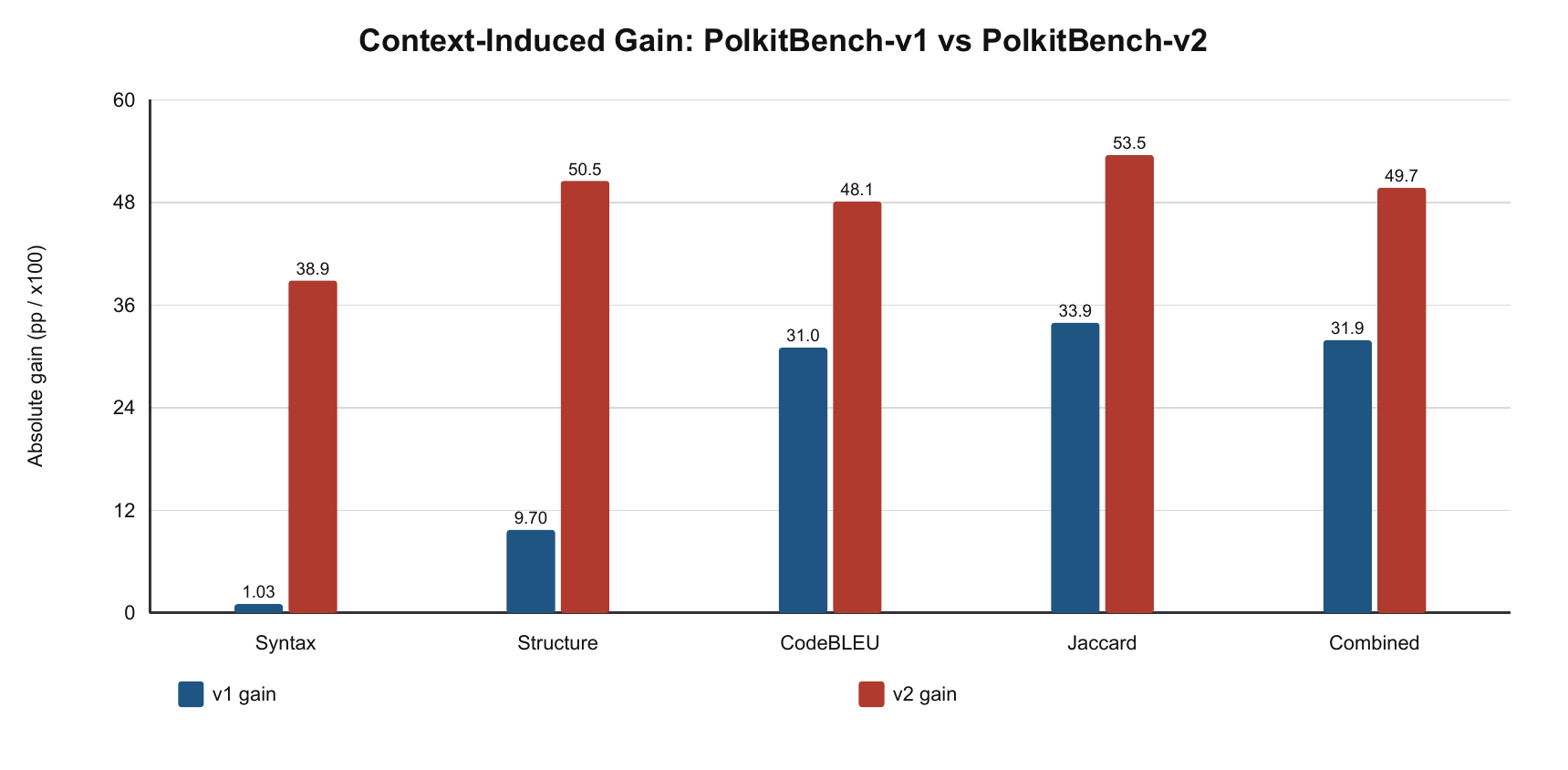}
\caption{Context-induced gain on PolkitBench-v1 vs.\ PolkitBench-v2 (GigaChat-10B). The gain is larger on the harder corpus on every metric.}
\label{fig:gain}
\end{figure}

\paragraph{Per-corpus context gain.} Within PolkitBench-v1 the context mode gains $+0.319$ on Combined relative to baseline; within PolkitBench-v2 the gain rises to $+0.497$ (Fig.~\ref{fig:gain}). The same pattern holds for Structure ($+9.7$\,pp vs.\ $+50.5$\,pp), Jaccard ($+0.339$ vs.\ $+0.535$) and CodeBLEU ($+0.310$ vs.\ $+0.481$). The larger gain on the harder corpus is consistent with the interpretation that the context mode is bounded by the model's intrinsic Polkit knowledge plus what $\mathrm{CM}$ supplies, while the baseline mode is bounded only by the former --- and the former degrades fast as queries leave the head of the distribution.

\paragraph{Strict Success interpretation.} Strict Success does \emph{not} follow the same monotone pattern, because it is the conjunction of four binary tests rather than a soft quality measure. On PolkitBench-v1 it decreases from $88.8$\,\% (baseline) to $77.1$\,\% (context); the context mode improves Structure, BLEU, Jaccard and Combined but is penalised on Strict Success. On PolkitBench-v2 the pattern reverses: $46.5$\,\% (baseline) $\to 60.0$\,\% (context). We therefore use Strict Success as an auxiliary indicator and rely on Syntax/Structure/CodeBLEU/Jaccard/Combined for cross-mode conclusions, as recommended in Section~\ref{sec:metrics}.

To pinpoint which of the four binary tests drives the v1 regression we decompose the $11.7$\,pp drop. The first three tests --- code extraction, syntactic validity and structural validity --- equal or exceed the baseline in the context mode: Syntax rises from $97.57\%$ to $98.60\%$ and Structure from $88.94\%$ to $98.64\%$. The regression is therefore not caused by malformed output. The remaining test --- zero hallucinated \texttt{action.id} --- is the dominant cause: the previous work~[1] reported $1{,}413$ invalid \texttt{action.id} instances on the v1 context run, and the corresponding figure on the GigaChat run used here is $1{,}075$ instances. With a mean of $\approx 1.0$ hallucinated identifier per affected record on v1, this accounts for $\approx 22\%$ of the $4{,}204$ records and almost exactly matches the $22.9\%$ Strict Success failure rate of the context mode. The v1 regression is therefore an artefact of the strict conjunction with the hallucination test, not a regression in structural or syntactic quality.

\paragraph{Residual hallucinations.} Even with full context, the absolute count of hallucinated \texttt{action.id} instances rises with corpus size: 1{,}075 on PolkitBench-v1 to 4{,}817 on PolkitBench-v2 in the context mode. Per-record this is comparable (0.26 vs.\ 0.48 per response on average); the absolute number nevertheless makes clear that residual hallucinations persist and are not eliminated by context injection. We discuss this in Section~\ref{sec:limits}.

\subsection{Per-Component Ablation on GigaChat-10B}
\label{sec:ablation_results}

\begin{figure}[htbp]
\centering
\includegraphics[width=0.95\textwidth]{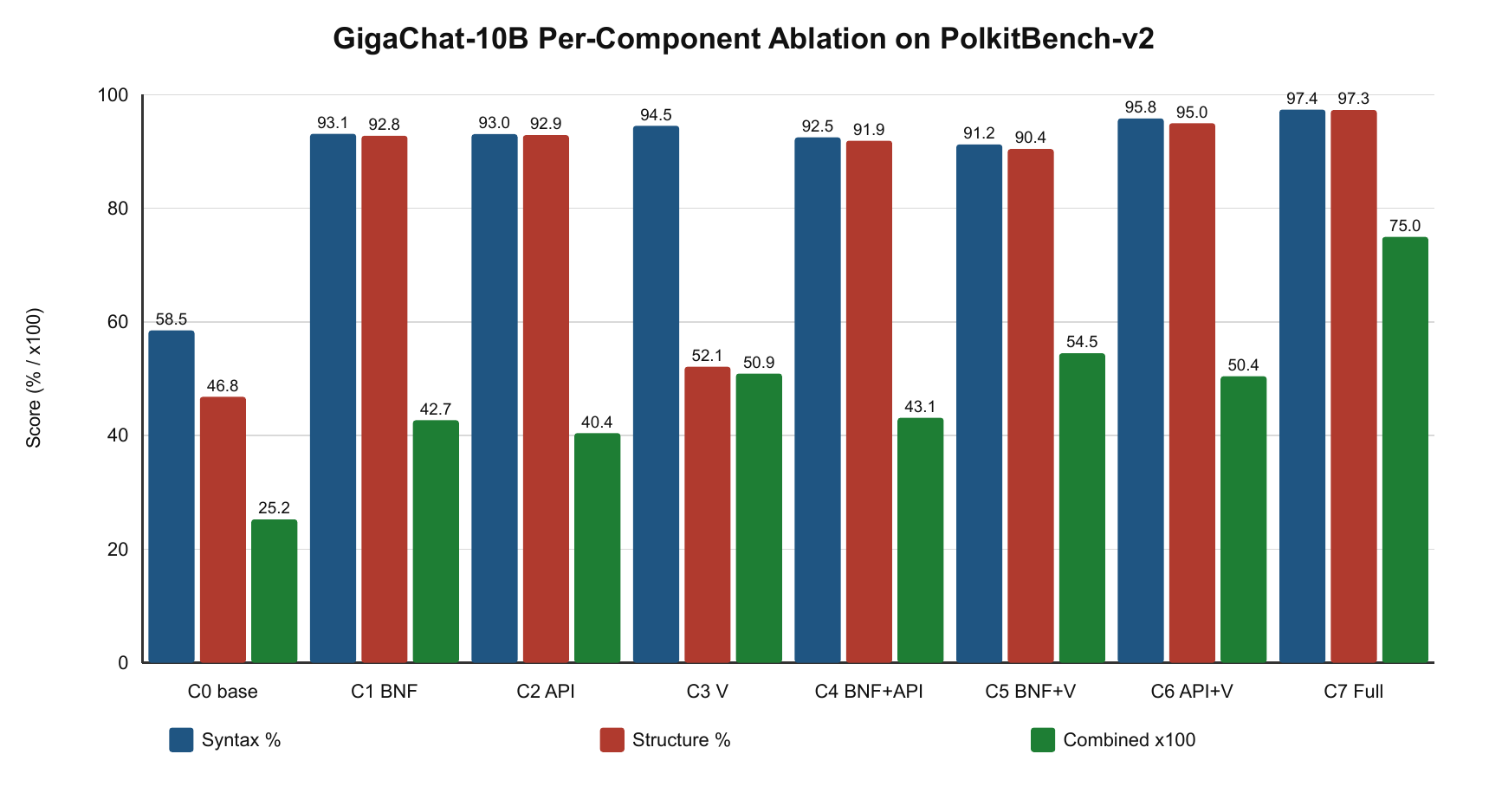}
\caption{Per-component ablation of structured context on GigaChat-10B and PolkitBench-v2. The full context $C_7$ is uniformly best; the strongest partial conditions $C_5$ and $C_6$ both contain the vocabulary.}
\label{fig:ablation}
\end{figure}

Table~\ref{tab:ablation} reports the full $2^3$ matrix on PolkitBench-v2; Fig.~\ref{fig:ablation} visualises the Strict Success and Combined Score columns.

\begin{table}[htbp]
\caption{Per-component ablation $C_0\ldots C_7$ on GigaChat-10B and PolkitBench-v2 ($N=10{,}073$). Bold marks the best value in each column. Values in brackets are $95\%$ Wilson confidence intervals for the binary metrics (Strict, Syntax, Structure); $95\%$ normal-theory CI half-widths for the continuous metrics (BLEU, Jaccard, Combined) are uniformly $\le 0.006$ across all conditions and are omitted from the table for compactness.}
\label{tab:ablation}
\begin{center}
\resizebox{\textwidth}{!}{%
\begin{tabular}{llrrrrrr}
\toprule
Cond. & Components & Strict [95\% CI] & Syntax [95\% CI] & Struct. [95\% CI] & BLEU & Jaccard & Comb. \\
\midrule
$C_0$ & ---                       & 46.53 [45.6,47.5] & 58.51 [57.5,59.5] & 46.81 [45.8,47.8] & 0.307 & 0.125 & 0.252 \\
$C_1$ & BNF                       & 15.73 [15.0,16.4] & 93.11 [92.6,93.6] & 92.77 [92.3,93.3] & 0.549 & 0.143 & 0.427 \\
$C_2$ & API                       & 25.20 [24.4,26.1] & 93.05 [92.5,93.5] & 92.87 [92.4,93.4] & 0.497 & 0.189 & 0.404 \\
$C_3$ & Vocab                     & 22.62 [21.8,23.4] & 94.52 [94.1,95.0] & 52.09 [51.1,53.1] & 0.571 & 0.365 & 0.509 \\
$C_4$ & BNF $+$ API               & 13.72 [13.1,14.4] & 92.47 [91.9,93.0] & 91.91 [91.4,92.4] & 0.552 & 0.149 & 0.431 \\
$C_5$ & BNF $+$ Vocab             & 50.49 [49.5,51.5] & 91.25 [90.7,91.8] & 90.42 [89.8,91.0] & 0.613 & 0.385 & 0.545 \\
$C_6$ & API $+$ Vocab             & 38.09 [37.2,39.0] & 95.80 [95.4,96.2] & 94.97 [94.5,95.4] & 0.568 & 0.357 & 0.504 \\
$C_7$ & BNF $+$ API $+$ Vocab     & \textbf{60.03} [59.1,61.0] & \textbf{97.37} [97.0,97.7] & \textbf{97.32} [97.0,97.6] & \textbf{0.788} & \textbf{0.660} & \textbf{0.750} \\
\bottomrule
\end{tabular}}
\end{center}
\end{table}

\paragraph{Finding 1: Full context is uniformly best.}
$C_7$ achieves the best value of every metric reported in Table~\ref{tab:ablation}. This contrasts with pilot results on different teacher models discussed in our prior work~[1], where full context occasionally lagged a strong partial configuration; on GigaChat-10B no such inversion is observed on PolkitBench-v2.

\paragraph{Finding 2: Among partials, $C_5$ and $C_6$ dominate.} Among the six partial conditions, $C_5$ (BNF $+$ Vocabulary) is best on Strict Success ($50.5$\,\%), CodeBLEU ($0.613$), Jaccard ($0.385$) and Combined ($0.545$). $C_6$ (API $+$ Vocabulary) is best on Syntax Valid ($95.8$\,\%) and Structure Valid ($95.0$\,\%). Both top partial conditions contain the vocabulary.

\paragraph{Finding 3: BNF alone is structure-without-semantics.} $C_1$ (BNF only) lifts Syntax from $58.5$\,\% to $93.1$\,\% and Structure from $46.8$\,\% to $92.8$\,\% --- a large gain on formal-correctness metrics. But it drops Strict Success to $15.7$\,\% and yields Combined $0.427$, only marginally above $C_0$. The grammar alone teaches the model to write formally-shaped rules but does not constrain the identifiers used inside them, which is what Strict Success and Jaccard penalise.

\paragraph{Finding 4: API alone is weaker than vocabulary alone.} $C_2$ (API only) is similar to $C_1$ on Syntax/Structure but slightly behind on Jaccard ($0.189$); $C_3$ (Vocabulary only) is the strongest single-component condition on Combined ($0.509$, vs.\ $0.427$ for $C_1$ and $0.404$ for $C_2$). The Structure score in $C_3$ collapses to $52.1$\,\%, however, because without BNF or API the model frequently emits rules outside the expected callback shape.

\paragraph{Finding 5: BNF $+$ API without vocabulary is the worst non-baseline partial.} $C_4$ scores $13.7$\,\% Strict Success and Jaccard $0.149$, both worse than several single-component conditions. The interaction is negative on semantic metrics: grammar and API jointly tighten the form of the rule while leaving the model free to invent identifiers, which the strict and Jaccard tests both penalise.

\paragraph{Finding 6: Vocabulary alone collapses structural validity.} $C_3$ (Vocabulary only) achieves Structure Valid $52.1\%$ --- far below all other contextual conditions and only marginally above the baseline ($46.8\%$) --- despite having the second-best partial Jaccard ($0.365$) and a competitive CodeBLEU ($0.571$). The interpretation is direct. The vocabulary $V$ specifies the legal string-valued identifiers but does not constrain the callback signature, the top-level \texttt{polkit.addRule(\ldots)} envelope or the use of \texttt{polkit.Result.*} return values. The model knows \emph{which} \texttt{action.id} values to use, but no longer how to wrap them, so it frequently emits rules with correct identifiers in an incorrect structural envelope. This asymmetry confirms that the three context components address \emph{different} failure modes: vocabulary fixes identifier choice, while BNF and API jointly fix structural correctness. It also explains the negative interaction observed in Finding~5 and underpins the recommendation R1 of Section~\ref{sec:discussion}: any subset of $\mathrm{CM}$ that omits BNF and API simultaneously is structurally fragile.

\paragraph{Finding 7: BNF and API have negative main effects on Strict Success.} Table~\ref{tab:effects} reports $E_{\text{BNF}}^{(\text{Strict})} = -1.79$ and $E_{\text{API}}^{(\text{Strict})} = -7.74$, both negative. This is the same mechanism that drives the v1 regression analysed in Section~\ref{sec:bc_results}: supplying BNF or API in isolation makes the model produce more structurally elaborate rules with more identifier occurrences per response, while leaving identifier choice unconstrained. The expanded surface area amplifies the probability of containing at least one out-of-vocabulary identifier and consequently zeroes Strict Success. Vocabulary has the opposite sign ($+17.52$) because it removes the very degree of freedom that BNF and API would otherwise enlarge. The negative main effects on Strict Success therefore do not contradict the positive main effects of the same factors on syntactic and structural validity ($+13.92$ and $+10.81$, $+22.33$ and $+24.66$ respectively): each factor is helpful on the dimension it targets and harmful on the dimension it does not, which is precisely why their joint use ($C_7$) dominates.

\paragraph{Factor decomposition.} Table~\ref{tab:effects} reports main effects~\eqref{eq:main_effect} and Shapley values per metric. The pattern is consistent across decompositions: Vocabulary has the largest semantic effect (Combined $+0.198$, Jaccard $+0.290$, Strict Success $+17.5$\,pp); API and BNF have the largest structural effects (Shapley on Structure: API $+24.7$, BNF $+22.3$, Vocabulary $+3.5$). The three components are complementary rather than substitutable: their factorial sum on Combined ($0.198 + 0.111 + 0.066 = 0.375$ in main-effect terms) is comparable to the $C_7 - C_0$ gap ($0.498$), with the residual accounted for by positive two-factor interactions, principally BNF $\times$ Vocabulary and API $\times$ Vocabulary.

\begin{table}[htbp]
\caption{Main effects $E_X^{(m)}$ (Eq.~\ref{eq:main_effect}) and Shapley values $\phi_X^{(m)}$ on GigaChat-10B and PolkitBench-v2. Main effect and Shapley value coincide for symmetric three-factor designs.}
\label{tab:effects}
\begin{center}
\setlength{\tabcolsep}{6pt}
\begin{tabular}{lrrrrrr}
\toprule
Factor & Strict & Syntax & Structure & BLEU & Jaccard & Combined \\
\midrule
BNF        & $-1.79$ & $+13.92$ & $+22.33$ & $+0.072$ & $-0.005$ & $+0.066$ \\
API        & $-7.74$ & $+10.81$ & $+24.66$ & $+0.045$ & $+0.018$ & $+0.046$ \\
Vocabulary & $+17.52$ & $+\phantom{0}9.21$ & $+\phantom{0}3.52$ & $+0.131$ & $+0.290$ & $+0.198$ \\
\bottomrule
\end{tabular}
\end{center}
\end{table}

\section{Discussion}
\label{sec:discussion}

\subsection{Practical Implications}

The combined evidence of Sections~\ref{sec:bc_results} and~\ref{sec:ablation_results} translates into three actionable recommendations for practitioners building a Text2DSL pipeline.

\noindent\textbf{(R1) Keep the full context.} The unique uniformly-best condition is $C_7$. There is no smaller subset of $\mathrm{CM}$ on GigaChat-10B that is competitive across all five primary metrics. Pruning $\mathrm{CM}$ to ``the one strongest piece'' is not supported.

\noindent\textbf{(R2) If the context must be reduced, retain the vocabulary.} The two best partial conditions ($C_5$, $C_6$) both contain the vocabulary; the worst non-baseline partial ($C_4 = \text{BNF} + \text{API}$) lacks it. Vocabulary contributes the largest semantic-quality effect and is the cheapest component to maintain (it is a flat list rather than a grammar or a specification document).

\noindent\textbf{(R3) Treat the distilled corpus as a renewable artefact, not a frozen one.} PolkitBench-v2 is bound to a specific version of $\mathrm{CM}$ via per-record provenance. When $\mathrm{CM}$ is updated --- a new \texttt{action.id} subsystem, a tightened grammar, a revised \texttt{polkit.Result} interpretation --- the procedure of Algorithm~\ref{alg:distill} can be re-run to produce an updated corpus that carries the new provenance, without manual labelling.

\subsection{Relation to Prior PolkitBench Results}

The previous work~[1] reported context-induced gains on PolkitBench-v1 with GigaChat-10B of $+18.9$\,pp Syntax, $+35.5$\,pp Structure and $+0.399$ Combined; the corresponding gains on PolkitBench-v2 reported here are $+38.9$\,pp Syntax, $+50.5$\,pp Structure and $+0.497$ Combined. The two sets of numbers are consistent in direction and consistent in the qualitative ordering of metric sensitivities. The larger absolute gains on PolkitBench-v2 reflect the larger headroom on a harder corpus rather than a change in the underlying effect: on every metric the context mode sits in roughly the same band (Syntax $\approx 97$\,\%, Structure $\approx 97$\,\%, Combined $\approx 0.75$--$0.80$) on both corpora, while baseline drops markedly.

The two corpora also serve different roles. PolkitBench-v1 remains useful as a tighter, smaller, manually-reviewable benchmark; PolkitBench-v2 is the larger and operationally more realistic corpus that exposes failure modes invisible on the smaller one, in particular the timeout mode ($183 \to 24$ for context in v1; $901 \to 161$ in v2) and the long-tail identifier hallucinations.

\subsection{Limitations and Threats to Validity}
\label{sec:limits}

We list five limitations bounding the scope of the conclusions.

\noindent\textbf{1. One evaluated model.} The ablation is reported on a single evaluated model (GigaChat-10B-A1.8B). The previous work~[1] showed that the direction of the full-context effect is consistent across two MoE models of different scale and provenance; the per-component ranking reported here may shift on dense transformers or on models substantially smaller or larger than 10B parameters. Replicating the $2^3$ matrix on Nemotron-3-Nano-30B-A3B and on additional dense models is the natural next step.

\noindent\textbf{2. One teacher.} PolkitBench-v2 is distilled by a single teacher (DeepSeek-V4-Flash). The corpus inherits whatever distributional biases that teacher has under the supplied $\mathrm{CM}$, even though every record is independently verified. A multi-teacher distillation would lower this risk.

\noindent\textbf{3. Residual identifier hallucinations.} The context mode on PolkitBench-v2 still emits $4{,}817$ instances of out-of-vocabulary \texttt{action.id} strings ($\approx 0.48$ per response on average). This means Strict Success $=0$ for roughly $40\%$ of the records where the hallucination test is the sole cause of failure. Two implementation paths for mitigation are immediate and compatible with the distillation method of Section~\ref{sec:method}; a full mitigation experiment is reserved for a follow-up paper.

\emph{(a) Post-hoc identifier rewrite.} Each emitted \texttt{action.id} candidate is matched against $V$; if absent, it is replaced by its nearest neighbour in $V$ under Levenshtein distance, subject to a similarity threshold below which the response is left unchanged rather than rewritten incorrectly. This mechanism is consistent with the hallucination patterns reported in the previous work~[1], where most fabricated identifiers differ from the closest valid value by a small suffix substitution (e.g. \texttt{disable-net} vs.\ \texttt{disable-wifi}, \texttt{package-update} vs.\ \texttt{system-update}) and would therefore be recovered by edit-distance rewriting.

\emph{(b) Grammar-constrained decoding restricted to $V$.} Combining context injection with constrained decoding~\cite{poesia2022synchromesh,geng2023gcd} so that the decoder accepts only members of $V$ in identifier positions provides a hard guarantee but requires decoder-level integration with the local serving stack. We regard~(a) as the cheaper first step and~(b) as the principled long-term solution.

\noindent\textbf{4. Strict Success as a binary indicator.} Strict Success is an integrated end-to-end indicator and aggregates four binary tests. It is intentionally pessimistic and should not be read as a soft quality metric; we use it as a deployment-readiness signal.

\noindent\textbf{5. Runtime validation in a containerised environment.} The \texttt{polkitd} + \texttt{pkcheck} pipeline models the production decision path but is not a substitute for the full diversity of system configurations encountered in deployed Linux distributions. Edge cases tied to specific D-Bus configurations or to the interaction of Polkit with other authorisation layers are outside the scope of this evaluation.

\section{Conclusion}
\label{sec:conclusion}
We have presented a context-aware distillation method that produces verified Text2DSL corpora bound to an explicit context model, and a two-tier validation pipeline that combines AST parsing with runtime acceptance by the production Polkit daemon. Applied to Polkit with DeepSeek-V4-Flash as teacher, the procedure yields PolkitBench-v2: 10{,}073 verified natural-language--to--rule pairs at 100.0\,\% AST validity and 99.7\,\% runtime acceptance. The per-component factorial ablation of structured context on GigaChat-10B-A1.8B identifies the full context as the unique uniformly-best condition, identifies the vocabulary as the strongest semantic-quality driver, and identifies BNF and API as the strongest syntactic and structural stabilisers. The cross-corpus comparison shows that the context mode degrades only marginally on the harder PolkitBench-v2 (Combined Score $0.801 \to 0.750$) while the baseline mode collapses on every metric ($0.482 \to 0.252$), supporting the interpretation of structured context as a load-bearing mechanism rather than a cosmetic improvement.

Three directions for follow-up are immediate: replicating the $2^3$ ablation matrix on a dense transformer evaluated model; combining context injection with grammar-constrained decoding to drive the residual identifier hallucination rate to zero; and extending the distillation procedure to a second policy DSL (SELinux type enforcement or OPA Rego) to assess the portability of the method.

\subsubsection*{Declaration of Generative AI Tools}

A teacher large language model (DeepSeek-V4-Flash) was used to generate the synthetic candidate rules for PolkitBench-v2 as described in Section~\ref{sec:distillation}. Every candidate was independently validated by the AST and runtime pipeline of Section~\ref{sec:method}; no record was admitted into the corpus on the strength of the teacher's output alone. The authors reviewed and edited the manuscript and take full responsibility for its content.

\subsubsection*{Declaration of Competing Interests}

The authors declare no competing financial interests or personal relationships that could have appeared to influence the work reported in this paper.

\subsubsection*{Data Availability}

The PolkitBench-v1 (4{,}204 pairs) and PolkitBench-v2 (10{,}073 pairs) corpora, the context model $\mathrm{CM}$, the AST and runtime validators, the per-condition evaluation outputs and the factorial-ablation run logs are released as part of the AI-Polkit research project.

%
% ---- Bibliography ----
%

\end{document}